\documentclass[11pt,a4paper]{article}
\usepackage[hyperref]{naaclhlt2018}
\usepackage[utf8]{inputenc}  
\usepackage{times}
\usepackage{microtype}
\usepackage{adjustbox}
\usepackage{latexsym}
\usepackage{url}
\usepackage{mathtools}
\usepackage{multirow}
\usepackage{subcaption}
\usepackage{xspace}
\usepackage{afterpage}
\usepackage{pdflscape}
\usepackage{array}
\usepackage{arydshln}
\usepackage{float}
\usepackage{dcolumn}
\usepackage{amssymb,amsmath}
\usepackage{rotating}
\usepackage{xcolor}
\usepackage{enumitem}
\usepackage{footmisc}
\definecolor{darkblue}{rgb}{0,0,0.5}

\usepackage{braket}

\newenvironment{squishlist}%
{\begin{list}{$\bullet$}
  { \setlength{\itemsep}{0.1pt}
     \setlength{\parsep}{2.3pt}
     \setlength{\topsep}{2.3pt}
     \setlength{\partopsep}{0.1pt}
     \setlength{\leftmargin}{1em}
     \setlength{\labelwidth}{1em}
     \setlength{\labelsep}{0.4em} } 
}{\end{list}
}
 
\usepackage{url}
\makeatletter  
\g@addto@macro{\UrlBreaks}{\UrlOrds}
\makeatother

\aclfinalcopy 

\newcommand{\ignore}[1]{}

\usepackage{xcolor}
\usepackage[normalem]{ulem}
\definecolor{darkgreen}{rgb}{0.0, 0.5, 0.0}
\def\OD#1{{\color{darkgreen}OD: \it #1}}
\def\ODdel#1{\bgroup\markoverwith{\textcolor{darkgreen}{\rule[0.5ex]{2pt}{1pt}}}\ULon{#1}}

\def\JN#1{{\color{blue}JN: \it #1}}
\def\JNdel#1{\bgroup\markoverwith{\textcolor{blue}{\rule[0.5ex]{2pt}{1pt}}}\ULon{#1}}

\def\VR#1{{\color{red}VR: \it #1}}
\def\VRdel#1{\bgroup\markoverwith{\textcolor{red}{\rule[0.5ex]{2pt}{1pt}}}\ULon{#1}}

\title{How to Increase Reliability of Human Evaluations for Generation}
\title{Experimental Setups to Increase Reliability of Human Evaluations for\\ \JN{Natural Language} Generation}
\title{How to Increase the Reliability of Human Evaluations\\ for Natural Language Generation}

\title{Increasing the Reliability of Human Evaluation\\ for Natural Language Generation by Experimental Design}

\title{RankME: Reliable Human Ratings for Natural Language Generation}

\author{Jekaterina Novikova, Ondřej Dušek and Verena Rieser\\
  Interaction Lab \\
 Heriot-Watt University \\
  Edinburgh, UK \\
  {\tt j.novikova, o.dusek, v.t.rieser@hw.ac.uk} }

\date{}

\begin{document}
\maketitle
\begin{abstract}
Human evaluation for natural language generation (NLG)  often suffers from inconsistent user ratings. 
While previous research tends to attribute this problem to individual user preferences, we show that the quality of human judgements can also be improved by experimental design. 
We present a novel \emph{rank-based magnitude estimation} method (RankME), which combines the use of continuous scales  and relative assessments. We show that RankME significantly improves the reliability and consistency  of human ratings compared to traditional evaluation methods.
In addition, we show that it is possible to evaluate NLG systems according to  multiple, distinct criteria, which is important for error analysis. 
 Finally, we demonstrate that RankME, in combination with Bayesian estimation of system quality,
is a cost-effective alternative for ranking multiple NLG systems. 

\end{abstract}

\section{Introduction}

Human judgement is the primary evaluation criterion for language generation tasks \cite{gkatzia:enlg2015}. 
However, limited effort has been made to improve the reliability of these subjective ratings 
 \cite{gatt2017survey}.
In this research, we systematically compare and analyse a wide range of alternative experimental designs for eliciting intrinsic user judgements for the task of comparing multiple systems.
We draw upon previous studies in language generation, e.g. \cite{Belz:2010,belz2011discrete,Siddharthan:2012:MagintuteEstimation}, as well as in the related field of machine translation (MT), e.g.\ \cite{Bojar:WMT16,bojar2017findings}.
In particular, we investigate the following challenges:

\noindent 
{\bf Distinct criteria:}
Traditionally, NLG outputs are evaluated according to different criteria, such as naturalness and informativeness \cite{gatt2017survey}.
Naturalness, also known as fluency or readability, targets the linguistic competence 
 of the text.  Informativeness, otherwise known as accuracy or adequacy, targets the  relevance and correctness of the output relative to the input specification. 
Ideally, we want to measure outputs of NLG systems with respect to these 
distinct criteria, especially for error analysis. For instance, one system may produce syntactically fluent output but misses important information, while another system, although being less fluent, may generate output that covers the meaning perfectly. 
Nevertheless, human judges  
often fail to distinguish between these different aspects, 
 which results in highly correlated scores, e.g.\ \cite{Novikova:EMNLP2017}.
 This is one of the reasons why some more recent research adds a general, overall quality criterion \cite{wen_stochastic_2015,wen:emnlp2015,manishina_automatic_2016, novikova2016crowd,Novikova:EMNLP2017}, or even uses only that \cite{SharmaHSSB16}.
 In the following, we show that discriminative ratings for different aspects can still be obtained, using distinctive task design.

\noindent {\bf Consistency:} 
Previous research has identified a high degree of 
inconsistency in human judgements of NLG outputs, where ratings often differ significantly ($p<0.001$) for the same utterance \cite{Walker:2007}.
While this might be attributed to individual preferences, e.g. \cite{Walker:2007,dethlefs2014cluster}, we also show that consistency (as measured by inter-annotator agreement) can be improved 
by different experimental setups, e.g.\ the use of continuous scales instead of discrete ones. Inconsistent user ratings are problematic in many ways, e.g.\ when developing metrics for 
automatic evaluation
\cite{dusek_referenceless_2017,Novikova:EMNLP2017}.

\noindent {\bf Relative vs.\ absolute assessment.} 
Intrinsic human evaluation methods are typically designed to \emph{assess} the quality of a system. However, they are frequently used to \emph{compare} the quality of different NLG systems, which is not necessarily appropriate. In the following, we show that {\em relative} assessment methods produce more consistent and more discriminative human ratings than direct assessment methods. 

In order to investigate these challenges, we compare several state-of-the-art NLG systems, which are 
evaluated by human crowd workers using a range of evaluation setups. We show that our newly introduced method, called \emph{rank-based magnitude estimation} (RankME), 
outperforms traditional evaluation methods. It combines advances suggested by previous research, such as continuous scales \cite{belz2011discrete}, magnitude estimation \cite{Siddharthan:2012} and relative assessment \cite{callison-burch_meta-_2007}. 
All code and data, as well as a more detailed description of the study setup are publicly available at: \url{https://github.com/jeknov/RankME}

\vspace{-0.1cm}
\section{Experimental Setup}
\vspace{-0.1cm}
 
 We were able to obtain outputs of 3 systems from the recent E2E NLG challenge \cite{Novikova:Sigdial2017}:\footnote{\label{fn:e2e-web}\url{http://www.macs.hw.ac.uk/InteractionLab/E2E}} the \emph{Sheffield NLP} system \cite{sheffield:E2E} and the \emph{Slug2Slug} system \cite{juraska_slug2slug:_2018}, as well as the outputs of the baseline \emph{TGen} system \cite{Dusek:ACL16}. 
We chose these systems in order to assess whether our methods can discriminate between outputs of different quality:
 Automatic metric scores, including BLEU, METEOR, etc., 
 indicate that the \textit{Slug2Slug} and \textit{TGen} systems show similar performance 
 while \textit{Sheffield}'s is further apart.%
 \footref{fn:e2e-web}
 

All three systems are based on the sequence-to-sequence (seq2seq) architecture with attention \cite{bahdanau_neural_2015}. \emph{Sheffield NLP} and \emph{TGen} both use this basic architecture with LSTM recurrent cells \cite{hochreiter_long_1997} and a beam search, \emph{TGen} further adds a reranker to penalize semantically invalid outputs.
\emph{Slug2Slug} is an ensemble of three seq2seq models with LSTM recurrent decoders. Two of them use LSTM recurrent encoders and one uses a convolutional encoder. A reranker checking for semantic validity selects among the outputs of all three models.

We use the first one hundred outputs for each system, and 
we collect human ratings from three independent crowd workers for each output 
using the CrowdFlower 
 platform. 
We use three different methods to collect human evaluation data: 6-point Likert scales, plain magnitude estimation ({\em plain ME}), and rank-based magnitude estimation ({\em RankME}). In a magnitude  estimation  (ME)  task~\cite{Bard:1996}, 
subjects  provide a relative rating of an experimental sentence to  a  reference  sentence, which is associated with a pre-set/fixed number. 
If  the  target  sentence  appears  twice  as  good  as  the  reference  sentence,  for  instance, subjects are to multiply the reference score by two; if it appears half as good, they should divide it in half, etc.
Note that ME implies the use of continuous scales, i.e.\ rating scales without numerical labels, similar to the visual analogue scales used by \citet{belz2011discrete} or direct assessment scales of \cite{graham_continuous_2013,bojar2017findings}, however, without given end-points.
\citet{Siddharthan:2012:MagintuteEstimation} have previously used  ME for evaluating readability of automatically generated texts.
RankME extends this idea by asking subjects to provide a relative ranking of 
{\em all} target sentences. 
%
Table \ref{tab:methods} provides a summary of methods and scales, and indicates whether relative ranking or direct assessment was used. 
\ignore{
We used the first one hundred outputs for each system and 
we collect human ratings from three independent crowd workers for each output 
using the CrowdFlower 
 platform.\footnote{\url{https://www.crowdflower.com/}} 
 }

\begin{table}[t]
\centering
\footnotesize
\begin{tabular}{|l||c|c||c|c|}
\hline
Method        & DA & RR & DS & CS \\ \hline \hline
Likert  & x  &    & x  &    \\ \hline
Plain ME & x  &    &    & x  \\ \hline
RankME  &    & x  &    & x  \\ \hline
\end{tabular}
\vspace{-0.2cm}
\caption{Three methods used to collect human evaluation data. Here, DA = direct assessment, RR = relative ranking, DS = discrete scale, CS = continuous scale.}
\label{tab:methods}
\vspace{-0.3cm}
\end{table}


\vspace{-0.1cm}
\section{Judgements of Multiple Criteria}\label{sec:criteria}
\vspace{-0.1cm}

\ignore{
Traditionally, human evaluation aims to assess the naturalness and informativeness of an automatically generated output \cite{gatt2017survey}.
Naturalness, also known as fluency or readability, targets 
the linguistic quality of the text. 
Informativeness, otherwise known as accuracy or adequacy, targets the 
relevance or correctness of the output relative to the input, showing how well the system reflects the \OD{input} content. Some more recent 
research adds a general, overall quality criterion \cite{wen:emnlp2015,wen_stochastic_2015,manishina_automatic_2016, novikova2016crowd,Novikova:EMNLP2017}, or even uses only that \cite{SharmaHSSB16}.
}
In our experiments, we collect ratings on the following criteria:

\begin{squishlist}
\item {\bf Informativeness}~(=\,adequacy): {\em Does the utterance  provide all the useful information from the meaning representation?} 

\item {\bf Naturalness}~(=\,fluency): {\em Could the utterance have been produced by a native speaker?}

\item {\bf Quality:} {\em How do you judge the overall quality of the utterance in terms of its grammatical correctness, fluency, adequacy and other important factors?} 
\end{squishlist}

In order to investigate whether judgements of these criteria are correlated, we compare two experimental setups: In {\em Setup~1}, crowd workers are shown the input meaning representation (MR) and the corresponding output of one of the NLG systems and are asked to evaluate the output with respect to all three aspects in one task. 
In {\em Setup~2}, these aspects are assessed separately, in individual tasks. Furthermore, when crowd workers are asked to assess naturalness, the MR is not shown to them since it is not relevant for the task.
Both setups utilise 
all three data collection methods -- Likert scales, plain ME and RankME. 

The results in Table~\ref{tab:cor} 
 show that scores are highly correlated for Setup 1. 
 This is in line with previous research in MT 
\cite{callison-burch_meta-_2007,koehn_statistical_2010}. 
 Separate collection (Setup 2), 
 however, decreases correlation between naturalness and quality, as well as naturalness and informativeness to very low levels, especially when using ME methods. Nevertheless, informativeness and quality are still highly correlated.
 We assume that this is due to the fact that raters see the MR in both cases.

To obtain more insight into informativeness ratings, we asked crowd workers to further distinguish informativeness in terms of added and missed information with respect to the original MR. Crowd workers were asked to select a checkbox for \emph{added information} if the output contained information not present in the given MR, or a checkbox for \emph{missed information} if the output missed some information from the MR. 
The results of Chi-squared test show that distributions of missed and added information are significantly different (p $<$ 0.01), i.e.\ systems add or delete information at different rates. Again, this information is valuable for error analysis. 
In addition, results in Table~\ref{tab:rank} show that
 assessing the amount of missed information indeed produces a different overall system ranking to added information.
 %
As such, it is worth considering missed information
as a separate criterion for evaluation. This can also be approximated automatically, as demonstrated by \citet{Wiseman:EMNLP17}.

\begin{table*}
\centering
\footnotesize
\makebox[0pt][c]{\parbox{1\textwidth}{%

    \begin{minipage}[b]{0.32\textwidth}\centering
    \begin{adjustbox}{max width=1\textwidth}
        \begin{tabular}{ll|l|l|}
\cline{3-4}
 &  & \multicolumn{1}{c|}{\textbf{Setup 1}} & \multicolumn{1}{c|}{\textbf{Setup 2}} \\ \cline{3-4} 
 &  & \multicolumn{2}{c|}{naturalness} \\ \hline \hline
\multicolumn{1}{|l|}{Likert} & \multirow{3}{*}{\rotatebox[origin=c]{90}{quality}} & 0.54* & -0.01 \\ \cline{1-1}\cline{3-4}
\multicolumn{1}{|l|}{Plain ME} &  & 0.44* & -0.03 \\ \cline{1-1}\cline{3-4}
\multicolumn{1}{|l|}{RankME} &  & 0.28* & -0.04 \\ \hline
\end{tabular}
\end{adjustbox}
    \end{minipage}    
    \hfill    
    \begin{minipage}[b]{0.32\hsize}\centering
    \begin{adjustbox}{max width=1\textwidth}
        \begin{tabular}{ll|l|l|}
\cline{3-4}
 &  & \multicolumn{1}{c|}{\textbf{Setup 1}} & \multicolumn{1}{c|}{\textbf{Setup 2}} \\ \cline{3-4} 
 &  & \multicolumn{2}{c|}{informativeness} \\ \hline \hline
\multicolumn{1}{|l|}{Likert} & \multirow{3}{*}{\rotatebox[origin=c]{90}{quality}} & 0.00\phantom{3} & 0.54* \\ \cline{1-1}\cline{3-4}
\multicolumn{1}{|l|}{Plain ME} &  & 0.48* & 0.71* \\ \cline{1-1}\cline{3-4}
\multicolumn{1}{|l|}{RankME} &  & 0.55* & 0.74* \\ \hline
\end{tabular}
\end{adjustbox}
    \end{minipage}    
    \hfill
    \begin{minipage}[b]{0.32\hsize}\centering
    \begin{adjustbox}{max width=1\textwidth}
        \begin{tabular}{ll|l|l|}
\cline{3-4}
 &  & \multicolumn{1}{c|}{\textbf{Setup 1}} & \multicolumn{1}{c|}{\textbf{Setup 2}} \\ \cline{3-4} 
 &  & \multicolumn{2}{c|}{naturalness} \\ \hline \hline
\multicolumn{1}{|l|}{Likert} & \multirow{3}{*}{\rotatebox[origin=c]{90}{inform.}} & 0.15* & -0.18* \\ \cline{1-1}\cline{3-4}
\multicolumn{1}{|l|}{Plain ME} &  & 0.03 & -0.07 \\ \cline{1-1}\cline{3-4}
\multicolumn{1}{|l|}{RankME} &  & 0.09 & -0.08 \\ \hline
\end{tabular}
\end{adjustbox}
    \end{minipage}%
}}
\vspace{-2mm}
\caption{Spearman correlation between ratings of naturalness and quality, collected using two different setups and three data collection methods -- Likert, plain ME and RankME. Here, ``*'' denotes $p<0.05$.}
\label{tab:cor}
\vspace{-0.2cm}
\end{table*}

\vspace{-0.1cm}
\section{Consistency and Use of Scales}
\label{sec:scales}
\vspace{-0.1cm}

\begin{table}[t]
\vspace{-1mm}
\centering
\footnotesize
\begin{tabular}{|l|l|r|r|}
\hline
\textbf{Method} & \textbf{Rating} & \multicolumn{1}{l|}{\textbf{Setup 1}} & \multicolumn{1}{l|}{\textbf{Setup 2}} \\ \hline \hline
\multirow{3}{*}{Likert} & naturalness & 0.07\phantom{*} & 0.12\phantom{*} \\ 
\cline{2-4}  & quality & 0.02\phantom{*} & 0.41* \\ 
\cline{2-4}  & informativeness & 0.93* & 0.78* \\ \hline
\multirow{3}{*}{Plain ME} & naturalness & -0.03\phantom{*} & 0.27* \\ \cline{2-4} 
 & quality & 0.22* & 0.60* \\ 
 \cline{2-4}  & informativeness & 0.59* & 0.79* \\\hline
\multirow{3}{*}{RankME} & naturalness & 0.11\phantom{*} & \textbf{0.42*} \\ \cline{2-4} 
 & quality & 0.10\phantom{*} & \textbf{0.68*} \\
 \cline{2-4}  & informativeness & 0.72* & \textbf{0.82*} \\ \hline
\end{tabular}
\vspace{-1mm}
\caption{ICC scores for human ratings of naturalness, informativeness and quality. ``*'' denotes $p<0.05$.}
\label{tab:icc}
\vspace{-2mm}
\end{table}

To assess consistency in human ratings, we calculate the intra-class correlation coefficient (ICC), which measures inter-observer reliability 
 for more than two raters~\cite{landis1977measurement}. 
In our experiments, we compare 
discrete Likert scales with continuous scales implemented via ME with respect to 
 the resulting reliability of collected human ratings. 
The results in Table~\ref{tab:icc} show that the use of ME significantly increases ICC levels for naturalness and quality. This effect is especially pronounced 
 for Setup~2 where ratings are collected separately. 
 Both plain ME and RankME methods show a significant increase in ICC, with the RankME method showing the highest ICC results. This difference is most apparent for naturalness, where RankME shows an ICC of 0.42 compared to plain ME's 0.27. 
 %
For informativeness, Likert scales already provide satisfactory agreement.
 
 In previous research, discrete, ordinal Likert scales are the dominant method of human evaluation for NLG, 
 although they may produce  results where statistical significance is overestimated \cite{gatt2017survey}. 
Recent studies show that continuous scales 
allow subjects to give more nuanced judgements \cite{belz2011discrete,graham_continuous_2013,bojar2017findings}. 
Moreover, raters 
were found to strongly prefer continuous scales over discrete ones \cite{belz2011discrete}. 
In addition to this previous work, our results also show that continuous scales significantly improve reliability of human ratings when implemented via ME. 

\vspace{-0.1cm}
\section{Ranking vs Direct Assessment}
\label{sec:rank}
\vspace{-0.15cm}

\begin{table}[t]
\centering
\footnotesize

\begin{tabular}{|l|l|}
\hline
\textbf{Ranking} & \textbf{Rating criterion \& method} \\ \hline\hline
\begin{tabular}[c]{@{}l@{}}1. Slug2Slug\\ 2. TGen\\ 3. Sheffield NLP\end{tabular} & \begin{tabular}[c]{@{}l@{}}Plain ME informativeness \\ RankME quality \\ TrueSkill quality \\ added information\end{tabular} \\\hline
\begin{tabular}[c]{@{}l@{}}1. TGen\\ 2. Slug2Slug\\ 3. Sheffield NLP\end{tabular} & \begin{tabular}[c]{@{}l@{}}missing information\end{tabular} \\\hline
\begin{tabular}[c]{@{}l@{}}1.--2. Slug2Slug\\\quad\quad + TGen\\ 3. Sheffield NLP\end{tabular} & \begin{tabular}[c]{@{}l@{}}Plain ME quality \\ RankME informativeness \\ TrueSkill informativeness \\ Likert quality \\ Likert informativeness \end{tabular} \\\hline
\begin{tabular}[c]{@{}l@{}}1.--2. Slug2Slug \\\quad\quad + Sheffield NLP\\ 3. TGen \end{tabular} & \begin{tabular}[c]{@{}l@{}}Likert naturalness \end{tabular} \\\hline
\begin{tabular}[c]{@{}l@{}}1.--3. Slug2Slug\\ \quad\quad + TGen \\ \quad\quad + Sheffield NLP\end{tabular} & \begin{tabular}[c]{@{}l@{}}Plain ME naturalness \\ RankME naturalness \\ TrueSkill naturalness \end{tabular} \\\hline
\end{tabular}

\vspace{-2mm}
\caption{Results of system ranking using different data collection methods with Setup 2 (different ranks are statistically significant with $p<0.05$).}
\label{tab:rank}
\vspace{-0.3cm}
\end{table}

Most data collection methods for evaluation, 
including Likert and plain ME, are designed to directly {\em assess} the quality of a system. 
 However, these methods are almost always used to {\em compare} multiple systems {\em relative} to each other. 
 Recently, the NLP evaluation literature has started to address this issue, mostly using binary comparisons, for example between the outputs of two MT 
 systems 
\cite{dras:2015,Bojar:WMT16}. 
In our experiments, Likert and plain ME are direct assessment (DA) methods, while RankME is a relative ranking (RR)-based method (see also Table~\ref{tab:methods}).
In order to directly compare DA and RR, we generated overall system rankings based on our different methods,
using pairwise bootstrap test at 95\% confidence level \cite{koehn_statistical_2004} to establish statistically significant differences.

The results in Table~\ref{tab:rank} show that
both plain ME and RankME methods produce similar rankings of NLG systems, which is in line with previous research in MT 
\cite{Bojar:WMT16}. 
It is also apparent that ME methods, by using a continuous scale, provide more distinctive overall rankings than Likert scales. For naturalness scores, no method results in clear system ratings, which possibly reflects in the low ICC of this criterion (cf.~Table~\ref{tab:icc}). RankME is the only method to provide a clear ranking with respect to overall utterance quality. However, its ranking of informativeness is less clear than that of plain ME, which might be due to 
the different results for missed and added information (see~Sec.~\ref{sec:scales}).
\ignore{However, we see diverging results between DA and RR assessment for ratings of naturalness, where only RankME is able to produce a distinctive ranking.
%
These results are in line with \cite{Belz:2010}, who compared a preference-based RR 
 to  a  standard  DA rating  scale  
 and  found  that  the  former  was  more sensitive to differences between systems, and less susceptible to variance between subjects.} 
In addition, the results in Table~\ref{tab:icc} 
show that RR, in combination with Setup 2, results in more consistent ratings than DA. 

\vspace{-0.1cm}
\subsection{Relative comparisons of many outputs}

While there are clear advantages to relative rank-based assessment, 
the amount of data needed for this approach grows quadratically with the number of systems to compare,
which is problematic with larger numbers of systems, e.g.\ in a shared task challenge. 
Data-efficient ranking algorithms, such as TrueSkill \cite{herbrich2007trueskill}, are therefore applied by recent MT evaluation studies~\cite{sakaguchi2014efficient,Bojar:WMT16} to produce overall system rankings based on a sample of binary comparisons. However, TrueSkill has not previously been used for evaluating NLG systems.
TrueSkill produces system rankings by gradually updating a Bayesian estimate of each system's capability  according to the ``surprisal'' of pairwise comparisons of individual system outputs. 
This way, fewer direct comparisons between systems are needed to establish their overall ranking. 
%
%
We computed system rankings using TrueSkill over comparisons collected via RankME and were able to show that it produces exactly the same system rankings for all three criteria as using RankME directly (see Table~\ref{tab:rank}), despite the fact that the comparisons are only used in a ``win-loss-tie'' fashion. 
This shows that RankME can be used with TrueSkill 
 to produce consistent rankings of a larger number of systems.

\ignore{
While there are clear advantages to relative assessment, this technique results in exponential data needs, which is especially problematic when a greater number of systems are compared, e.g. as part of shared challenges. 
More efficient ranking algorithms, such as TrueSkill~\cite{herbrich2007trueskill}, are therefore favoured in recent MT evaluation studies~\cite{Bojar:WMT16,sakaguchi2014efficient}, but was not previously used for evaluating NLG systems.

We used a rank-based TrueSkill algorithm to process quality and naturalness scores collected using a RR method. The results show that it provides similar results as RankME method with all possible combination between systems. For example, with quality scores, rank-based TrueSkill resulted in three clusters ordered the same way as the first ranking in Table~\ref{tab:rank}. With naturalness scores, however, rank-based TrueSkill was not able to differentiate between systems and put all of them into one cluster. These results suggest that rank-based TrueSkill can be used for evaluating NLG systems, the same as it is used in MT. However, a lot of data needs to be collected in order to produce significant differences among systems.

Another flaw of rank-based TrueSkill is that it loses a lot of information about actual quality that is provided by RankME scores. This additional information could help to create better/narrower clusters with the need of less data. Use of score-based TrueSkill~\cite{guo2012score} is a potential way to overcome this problem by combining advantages of RR and DA approaches, and this is a direction of our future research.
} 

\vspace{-0.1cm}
\section{Conclusion and Discussion}
\vspace{-0.1cm}

In this paper, we demonstrate that the experimental design has a significant impact on the reliability as well as the outcomes of human evaluation studies for natural language generation.  
We first show that correlation effects between different evaluation criteria can be minimised by eliciting them separately.
Furthermore, we introduce RankME, which combines relative rankings and magnitude estimation (with continuous scales), and demonstrate that this method results in better agreement amongst raters and more discriminative results.
Finally, our results suggest that TrueSkill is a cost-effective alternative for producing overall relative rankings of multiple systems.
This framework has the potential to not only significantly influence how NLG evaluation studies are run, but also produce more reliable data for further processing, e.g. for developing more accurate automatic evaluation metrics, which we are currently lacking, e.g. \cite{Novikova:EMNLP2017}.

In current work, we test RankME with a wider range of systems ({\em under submission}).
We also plan to investigate how this method transfers to related tasks, such as evaluating open-domain dialogue responses, e.g. \cite{Lowe:acl2017}.
In addition, we aim to investigate additional NLG evaluation methods, such as extrinsic task contributions, e.g.\ \cite{Rieser:IEEE14,Gkatzia:acl16}. 

\ignore{
In this work, we focus on intrinsic human evaluation methods, i.e. methods which measure the performance of a system without reference to its real-world use. Extrinsic evaluation metrics, such as NLG's contribution to task success, e.g.\ \cite{Rieser:IEEE14,Gkatzia:acl16}, or change in user perceptions \cite{Siddharthan:2012},
are in general considered to be desirable \cite{gatt2017survey}. Nonetheless, they are less efficient to implement. To the authors' knowledge, it is still to be shown that results from intrinsic human evaluation correlate with extrinsic measures. 

Finally, our study relies on crowd workers only. Previous research in NLG suggests that experts rate output differently to naive users \cite{belz2009system}. Whereas for MT translation .... \VR{ToDo: check MT literature.}
\OD{Bojar (WMT) 2013 shows that RR by experts has higher IAA than RR by crowd workers. Bojar (WMT) 2016 says DA/sliders done by crowd workers correlate strongly with RR+TrueSkill done by experts. This may be because that kind of DA is designed to be foolproof (there are sanity checks in place: people asked to rate the same thing twice, checks if bad outputs are indeed rated worse than human references).}
}

\section*{Acknowledgements}
 This research received funding from the EPSRC projects  DILiGENt (EP/M005429/1) and  MaDrIgAL (EP/N017536/1). The Titan Xp used for this research was donated by the NVIDIA Corporation.
 \vspace{-0.5cm}
\normalsize

\bibliography{naaclhlt2018}
\bibliographystyle{acl_natbib}

\end{document}